\documentclass[11pt]{article}

\usepackage[preprint]{acl}

\usepackage{times}
\usepackage{latexsym}

\usepackage[T1]{fontenc}

\usepackage[utf8]{inputenc}

\usepackage{microtype}

\usepackage{inconsolata}

\usepackage{graphicx}
\usepackage[table]{xcolor}
\usepackage{booktabs}
\usepackage{enumitem}
\usepackage{amsmath}
\usepackage{amssymb}
\usepackage{makecell}
\usepackage{multirow}
\usepackage{tabularx}
\usepackage[most]{tcolorbox}

\definecolor{paleaqua}{RGB}{214,238,238}

\title{HyGRAIL: Cost-Aware and Evidence-Grounded Scientific Hypothesis Discovery over Knowledge Graphs}

\author{
\textbf{Yihang Sun}\textsuperscript{1} \quad
\textbf{Zhihan Zhu}\textsuperscript{1,\textdagger} \quad
\textbf{Zhiyuan Jiang}\textsuperscript{1,\textdagger} \\
\textbf{Jingyi Ge}\textsuperscript{1,\textdagger} \quad
\textbf{Zixuan Li}\textsuperscript{1,\textdagger} \quad
\textbf{Jiaxuan You}\textsuperscript{1} \\
\textsuperscript{1}University of Illinois Urbana-Champaign \\
}

\begin{document}
\maketitle
\begingroup
\renewcommand{\thefootnote}{}
\footnotetext{\textsuperscript{\textdagger} Work done while interning at the University of Illinois Urbana-Champaign.}
\addtocounter{footnote}{-1}
\endgroup
\begin{abstract}
Scientific knowledge graphs organize entities and relations extracted from scientific literature, but they remain inherently incomplete. 
Missing typed links in such graphs can therefore represent plausible scientific hypotheses, such as unexplored associations between materials and applications. 
However, scientific hypothesis discovery is challenging because true discoveries are extremely sparse among typed candidate pairs: graph neural networks (GNNs) are efficient but unreliable for ambiguous cases, while large language models (LLMs) are knowledgeable but too costly to apply exhaustively and are not naturally grounded in graph structures. 
We propose \textbf{HyGRAIL}, a cost-aware and evidence-grounded framework that combines heterogeneous GNN triage with LLM-based hypothesis review. 
HyGRAIL first uses a GNN to score candidate hypotheses and identify a validation-calibrated ambiguous region, routing only graph-uncertain cases to LLM review. 
For each routed hypothesis, HyGRAIL retrieves node-level associations and multi-hop relational paths from the KG, then converts this structured evidence into natural language through template-based or LLM-based naturalization. 
An LLM review agent finally judges each hard hypothesis using the naturalized evidence and validation-selected decision criteria. 
On MatKG, HyGRAIL achieves the best F1 score of $0.429$, improving over the strongest prior baseline by $0.242$ F1 points and over the GNN-only baseline by $0.322$.
Meanwhile, GNN triage reduces the LLM call rate by $54.36\%$ on average.
Ablation studies further show that retrieved graph evidence is crucial for reliable hypothesis verification and that compact, two-sided evidence is more effective than simply increasing retrieval quantity.
\end{abstract}

\section{Introduction}

Scientific progress increasingly depends on the ability to synthesize knowledge scattered across rapidly growing scientific literature. 
Knowledge graphs (KGs) provide a natural substrate for this goal by representing domain entities and their typed relations in a structured form. 
Such graphs have been constructed across a wide range of scientific domains, including materials science, biomedicine, drug discovery, and scholarly knowledge organization~\citep{venugopal2024matkg,kilicoglu2012semmeddb,himmelstein2017hetionet,jaradeh2019orkg,zhang2025ikraph}. 
For example, MatKG~\citep{venugopal2024matkg} encodes materials-science entities such as chemicals, properties, applications, synthesis methods, characterization methods, and descriptors, together with typed associations extracted from scientific literature. 
In incomplete scientific KGs, many scientifically meaningful associations may be absent simply because they have not yet been reported, extracted, or connected in the graph. 
This naturally motivates a link-prediction view of scientific discovery: a missing typed link between two entities can be interpreted as a candidate scientific hypothesis to be verified~\citep{swanson1986undiscovered,sosa2020literature,sybrandt2020agatha,borrego2025research}. 
For instance, a missing \texttt{CHM--APL} link in MatKG may suggest that a chemical has unexplored potential for a particular application.

This formulation turns scientific discovery into a typed link prediction problem~\citep{sosa2020literature,sybrandt2020agatha,borrego2025research}, but in our setting discovery-relevant links are extremely sparse among all plausible typed node pairs, as shown in Table~\ref{tab:matkg_sparsity}. 
For a hypothesis type such as \texttt{CHM--APL}, the candidate space grows with the Cartesian product between chemicals and applications, while only a tiny fraction of these pairs are supported by existing scientific evidence. 
Table~\ref{tab:matkg_sparsity} illustrates this sparsity for the seven MatKG hypothesis types considered in our study. 
This extreme imbalance makes scientific hypothesis discovery both statistically difficult and practically high-stakes: false positives may waste downstream expert attention or experimental resources, while false negatives may overlook promising discoveries.

\begin{table}[t]
\centering
\small
\setlength{\tabcolsep}{12pt}
\begin{tabular}{@{}lccc@{}}
\toprule
\textbf{Type} & $\mathbf{P}$ $(10^5)$ & $\mathbf{C}$ $(10^8)$ & $\boldsymbol{\rho}$ $(10^{-3})$ \\
\midrule
CHM--APL & 1.06 & 1.91 & 0.556 \\
CHM--PRO & 3.40 & 4.27 & 0.796 \\
PRO--APL & 1.18 & 3.23 & 0.367 \\
SMT--PRO & 0.757 & 1.82 & 0.417 \\
CHM--SMT & 0.831 & 1.07 & 0.774 \\
CHM--DSC & 1.66 & 0.927 & 1.79 \\
CHM--CMT & 2.10 & 1.96 & 1.07 \\
\bottomrule
\end{tabular}
\vspace{-5pt}
\caption{
\textbf{Discovery-oriented hypothesis links are extremely sparse in MatKG.}
$P$ denotes positive links, $C$ denotes candidate pairs, and $\rho$ denotes density.
}
\label{tab:matkg_sparsity}
\vspace{-15pt}
\end{table}

Graph-based models are natural candidates for this problem because they can exploit the relational structure of scientific KGs. 
Classical KG embedding methods and relational GNNs have been widely used for link prediction over multi-relational graphs~\citep{bordes2013transe,yang2015distmult,schlichtkrull2018rgcn}, while heterogeneous GNNs further model node and edge type information through mechanisms such as meta-path attention or type-specific transformations~\citep{wang2019han,hu2020hgt}. 
However, graph-only models primarily rely on observed topology and relation statistics. 
In sparse scientific discovery settings, this often leads to a large ambiguous region in the predicted score distribution: positive and negative hypotheses may be separable at the two extremes, but heavily mixed in the middle~\citep{wang2021beconfident,huang2023conformalized,zhao2024conformalized}. 
As a result, even strong GNNs can serve as efficient scoring models but remain unreliable as final arbiters for difficult scientific hypotheses.

LLMs offer a complementary strength. 
Recent studies have explored their potential for scientific synthesis, hypothesis generation, and biomedical or chemistry-oriented discovery~\citep{zheng2023scientificllm,zhou2024hypothesisgeneration,qi2024biomedicalhypothesis,yang2024moosechem}. 
Their broad parametric knowledge and natural-language reasoning ability make them appealing as scientific hypothesis reviewers. 
Yet applying LLMs directly to all candidate links in a KG is computationally prohibitive, especially when the vast majority of candidates are likely negative. 
Moreover, LLMs are not naturally designed to consume raw graph structures~\citep{li2024graphllms,zhang2024largegraphmodels,cao2025unlocking}. 
Unguided LLM review may also rely on incomplete parametric memory rather than explicit evidence, motivating retrieval-based grounding~\citep{lewis2020rag,guu2020realm,karpukhin2020dpr}. 
Retrieval-augmented generation alleviates such grounding issues by conditioning LLMs on external knowledge~\citep{lewis2020rag}, and recent KG-guided RAG methods further show that graph relations can help organize retrieved evidence~\citep{zhu2025kg2rag}. 
However, existing KG-RAG work primarily targets question answering or text generation, whereas scientific hypothesis discovery requires evidence-grounded verification of typed candidate links.

We propose \textbf{HyGRAIL}, a hybrid framework for cost-aware and evidence-grounded scientific hypothesis discovery over heterogeneous KGs. 
HyGRAIL uses a heterogeneous GNN as a lightweight triage model: after training on the observed training graph, it identifies a validation-calibrated ambiguous score interval where graph-only predictions are unreliable. 
Only hypotheses falling into this hard region are routed to an LLM review agent, while easy hypotheses are handled directly by the GNN. 
For each routed hypothesis, HyGRAIL retrieves hypothesis-relevant graph evidence, including node-level associations and multi-hop relational paths, using both raw edge support and normalized relation importance. 
The structured evidence is then converted into natural language through either template-based or LLM-based naturalization, allowing the review agent to reason over scientific evidence without directly manipulating KG triples. 
Finally, the agent accepts a hypothesis only when both its binary decision and confidence score satisfy validation-selected criteria. 
This design combines the scalability of GNNs with the reasoning ability of LLMs while keeping LLM calls focused on the cases where they are most needed.

We summarize our contributions as follows:
\begin{itemize}[leftmargin=1.2em, itemsep=0pt, topsep=2pt, parsep=0pt, partopsep=0pt]
    \item We formulate scientific discovery as cost-aware, evidence-grounded hypothesis verification over incomplete weighted heterogeneous KGs.
    \item We introduce \textbf{HyGRAIL}, a hybrid GNN--LLM framework that uses validation-calibrated GNN score distributions to route only ambiguous hypotheses to LLM review.
    \item We design hypothesis-guided graph evidence retrieval and naturalization methods that transform node-level and multi-hop KG evidence into LLM-readable scientific evidence.
   \item We evaluate HyGRAIL across multiple GNNs and LLMs, achieving the best F1 score of $0.429$ and reducing the LLM call rate by $54.36\%$ on average. Ablations further show that retrieved graph evidence and two-sided endpoint context are crucial for reliable hypothesis verification.
\end{itemize}
\begin{figure*}[t]
    \vspace{-15pt}
    \centering
    \includegraphics[width=\textwidth]{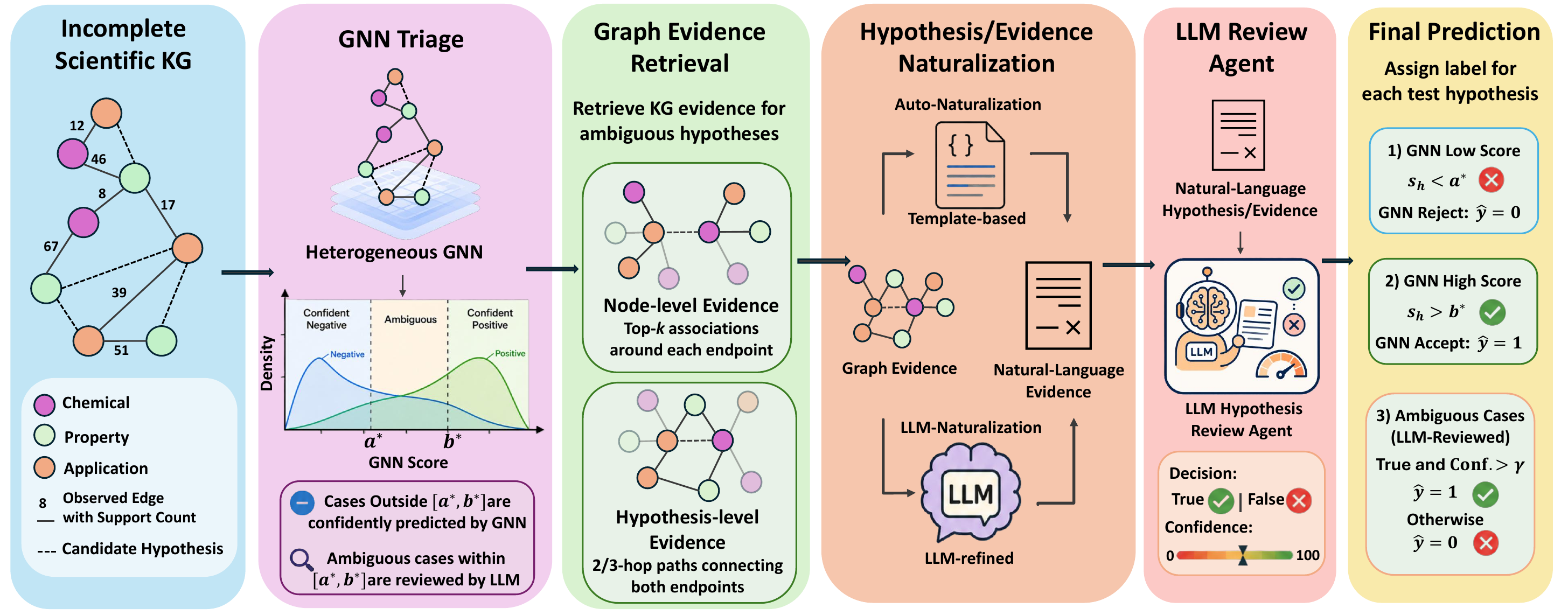}
    \vspace{-5pt}
    \caption{
    Overview of \textbf{HyGRAIL}. 
    HyGRAIL uses a heterogeneous GNN to triage candidate hypotheses, retrieves graph evidence for ambiguous hypotheses, naturalizes structured evidence into language, and asks an LLM review agent to make evidence-grounded decisions.
    }
    \label{fig:overview}
    \vspace{-15pt}
\end{figure*}

\section{Task Formulation}
\label{sec:task}

\subsection{Scientific Knowledge Graphs}

We consider a scientific knowledge graph as a weighted heterogeneous graph $\mathcal{G} = (\mathcal{V}, \mathcal{E}, \phi, \psi, c)$, where $\mathcal{V}$ is the set of entities, $\mathcal{E}$ is the set of observed edges, $\phi: \mathcal{V} \rightarrow \mathcal{T}_V$ maps each node to a node type, $\psi: \mathcal{E} \rightarrow \mathcal{T}_E$ maps each edge to an edge type, and $c(e) \in \mathbb{N}^{+}$ denotes the support count of edge $e$~\citep{venugopal2024matkg,himmelstein2017hetionet,jaradeh2019orkg}. 
The support count records how many scientific papers support the association represented by the edge~\citep{venugopal2024matkg}. 
We treat edges as undirected typed associations, where the semantics of an edge are determined by the types of its endpoints.

\subsection{Hypotheses as Typed Missing Links}

Not all edge types in a scientific KG are necessarily discovery targets. 
We define a set of hypothesis edge types $\mathcal{R}_{H} \subseteq \mathcal{T}_E$, where each $r \in \mathcal{R}_{H}$ specifies a type of scientific association to be discovered~\citep{himmelstein2017hetionet,sybrandt2020agatha}. 
A candidate hypothesis is a typed node pair $h=(u,v,r)$, where $u, v \in \mathcal{V}$ and the endpoint types of $u$ and $v$ match the relation type $r$. 
For example, a \texttt{CHM--APL} hypothesis states that a chemical may be useful for a particular application. 
The hypothesis space for relation type $r$ is defined as
\[
\mathcal{H}_{r} = 
\{(u,v,r) \mid \phi(u)=t_u(r),\ \phi(v)=t_v(r)\},
\]
where $t_u(r)$ and $t_v(r)$ denote the endpoint node types associated with relation type $r$. 
The full hypothesis space is $\mathcal{H}=\bigcup_{r\in\mathcal{R}_{H}}\mathcal{H}_{r}$.

As shown in Table~\ref{tab:matkg_sparsity}, these hypothesis spaces are highly sparse: only a small fraction of typed candidate pairs are observed as positive links. 
This sparsity makes exhaustive manual or LLM-based review impractical, and also makes graph-only prediction unreliable for ambiguous hypotheses.

\subsection{Closed-World Evaluation and Candidate Construction}

For evaluation, observed edges whose types belong to $\mathcal{R}_{H}$ are treated as positive hypotheses.
Unlinked typed node pairs sampled from the same hypothesis space are treated as negative hypotheses under a closed-world evaluation protocol~\citep{bordes2013transe,yang2015distmult,trouillon2016complex,kotnis2017negative}.
Let $\mathcal{E}_{H}$ denote observed hypothesis edges and $\mathcal{U}_{H}$ denote sampled unlinked candidates.
Each evaluated hypothesis $h=(u,v,r)$ receives label $y_h=1$ if $h \in \mathcal{E}_{H}$ and $y_h=0$ if $h \in \mathcal{U}_{H}$.
This protocol follows standard practice in link prediction~\citep{bordes2013transe,yang2015distmult,trouillon2016complex}, but we emphasize that an unobserved edge does not necessarily correspond to a scientifically false hypothesis~\citep{kotnis2017negative,kazemi2018simple}. 
Thus, negative labels are closed-world evaluation labels rather than definitive scientific invalidity.
In our benchmark, held-out observed edges serve as ground-truth positives for evaluation; in practical deployment, newly accepted unobserved links should be treated as candidate hypotheses for further expert or experimental validation.
\section{Method}
\label{sec:method}

\noindent\textbf{Overview.}
Given a scientific KG $\mathcal{G}$ and a set of candidate hypotheses $\mathcal{H}$, where each candidate hypothesis is a typed node pair $h=(u,v,r)$ defined in Section~\ref{sec:task}, our goal is to predict whether $h$ corresponds to a valid scientific association. 
As illustrated in Figure~\ref{fig:overview}, HyGRAIL consists of four components. 
(1) It trains a heterogeneous GNN to score hypotheses and identify an ambiguous score region where graph-only predictions are unreliable. 
(2) For hypotheses in this region, it retrieves hypothesis-relevant graph evidence, including node-level associations and multi-hop relational paths. 
(3) It converts structured evidence into natural language through either template-based or LLM-based naturalization. 
(4) It asks an LLM hypothesis review agent to judge each hard hypothesis using the naturalized evidence and validation-selected decision criteria. 
This design combines scalable graph representation learning~\citep{schlichtkrull2018rgcn,wang2019han,hu2020hgt} with evidence-grounded LLM reasoning~\citep{lewis2020rag,guu2020realm,karpukhin2020dpr}.

\subsection{GNN-based Hypothesis Triage}
\label{sec:gnn_triage}

HyGRAIL first uses a heterogeneous GNN as a lightweight triage model. 
Heterogeneous and relational GNNs are well suited for scientific KGs because they can model typed nodes and typed relations through relation-specific transformations, attention, or message passing~\citep{schlichtkrull2018rgcn,wang2019han,hu2020hgt}. 
For a candidate hypothesis $h=(u,v,r)$, the GNN encoder computes node representations and a relation-specific decoder outputs a score $s_h=f_{\theta}(u,v,r)$, where larger $s_h$ indicates stronger graph-based support. 
The GNN is trained on the training graph using observed hypothesis edges as positives and sampled unlinked typed pairs as negatives~\citep{bordes2013transe,yang2015distmult,trouillon2016complex,schlichtkrull2018rgcn}.

\begin{figure}[t]
    \centering
    \includegraphics[width=0.92\linewidth]{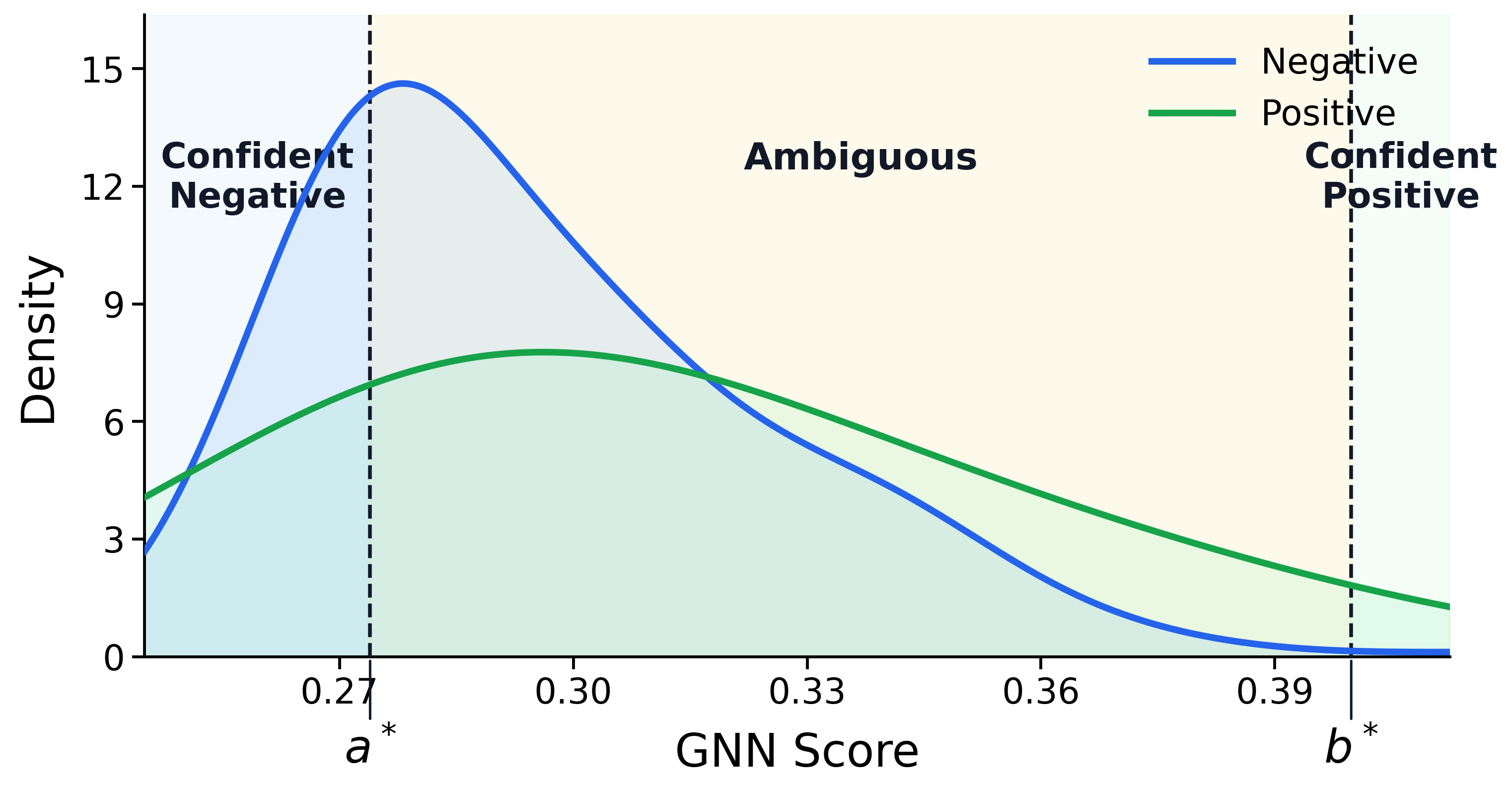}
    \vspace{-5pt}
    \caption{
    Validation-calibrated ambiguous region for GNN-based hypothesis triage.
This figure shows the HGT results on \texttt{PRO--APL} hypotheses.
HyGRAIL directly classifies confident low-score and high-score hypotheses, and routes only ambiguous hypotheses to LLM review.
    }
    \label{fig:score_distribution}
    \vspace{-15pt}
\end{figure}

HyGRAIL uses validation scores to decide where LLM review is needed. 
For a lower threshold $a$, we define the negative purity below $a$ as the fraction of validation hypotheses with $s_h<a$ whose labels are negative. 
Similarly, for an upper threshold $b$, we define the positive purity above $b$ as the fraction of validation hypotheses with $s_h>b$ whose labels are positive. 
Given two preset purity parameters $m$ and $n$, chosen according to the positive--negative ratio of each KG, HyGRAIL selects the smallest interval $[a^\ast,b^\ast]$ such that the negative purity below $a^\ast$ is at least $m$ and the positive purity above $b^\ast$ is at least $n$. 
Here, ``smallest'' means that the interval routes the fewest validation hypotheses to the LLM among all feasible intervals.

At inference time, HyGRAIL predicts negative if $s_h<a^\ast$, predicts positive if $s_h>b^\ast$, and routes $h$ to the LLM review agent only when $a^\ast \leq s_h \leq b^\ast$. 
This triage mechanism reserves costly LLM inference for hypotheses that graph-only models cannot confidently resolve.

\subsection{Hypothesis-Guided Graph Evidence Retrieval}
\label{sec:evidence_retrieval}

For each routed hypothesis $h=(u,v,r)$, HyGRAIL retrieves graph evidence from the KG before invoking the LLM. 
Unlike standard text retrieval, the retrieved evidence is structured, typed, and relation-dependent. 
HyGRAIL retrieves two complementary kinds of evidence: node-level evidence and hypothesis-level evidence.

\paragraph{Node-level evidence.}
For each endpoint node $x \in \{u,v\}$, HyGRAIL retrieves local associations that characterize the scientific context of $x$. 
For each node type $t$, we define an evidence edge type set $\mathcal{R}_{\mathrm{evi}}(t)$, which contains edge types considered informative for describing nodes of type $t$. 
For example, evidence for a chemical node may include its associated properties or applications, while evidence for an application node may include chemicals or properties associated with that application.

Let $\mathcal{N}_{r'}(x)$ denote the set of observed edges incident to $x$ whose edge type is $r' \in \mathcal{R}_{\mathrm{evi}}(\phi(x))$. 
For each evidence edge $e \in \mathcal{N}_{r'}(x)$, HyGRAIL uses both its raw support count $c(e)$ and its normalized weight $w(e)=c(e)/\sum_{e' \in \mathcal{N}_{r'}(x)}c(e')$. 
The support count captures how well established an association is in the literature, while the normalized weight captures how distinctive the association is among edges of the same type for node $x$.

To combine these two signals, we define
\[
\mathrm{NormCount}(e)=\frac{\log(1+c(e))}{\max_{e' \in \mathcal{N}_{r'}(x)}\log(1+c(e'))},
\]
which normalizes absolute support counts within the same evidence edge type. 
HyGRAIL then ranks candidate evidence edges by $\mathrm{EviScore}(e)=\alpha \cdot \mathrm{NormCount}(e)+(1-\alpha)\cdot w(e)$, where $\alpha$ balances absolute literature support and relative distinctiveness. 
The top-$k$ edges are retained as node-level evidence.

\paragraph{Hypothesis-level evidence.}
Node-level evidence describes each endpoint independently, whereas a scientific hypothesis concerns the relation between two endpoints. 
Therefore, HyGRAIL also retrieves short paths connecting $u$ and $v$, focusing on 2-hop and 3-hop paths as hypothesis-level relational evidence~\citep{lao2010relational,neelakantan2015compositional}. 
These paths provide structured signals about how the two endpoint entities are connected through intermediate scientific entities.

\subsection{Graph Evidence Naturalization}
\label{sec:naturalization}

Although LLMs can reason effectively over natural language, raw KG evidence is represented as typed edges and paths. 
Directly providing such structured triples to an LLM may force the model to interpret graph notation rather than evaluate the scientific hypothesis. 
Therefore, HyGRAIL naturalizes retrieved graph evidence into concise natural-language statements before review.

We consider two naturalization strategies.

\paragraph{Auto-Naturalization.}
Auto-Naturalization uses deterministic templates to convert graph evidence into natural language. 
For each hypothesis type and evidence type, we design templates that describe the scientific meaning of the corresponding edge or path. 
To preserve quantitative information, each evidence item is categorized by whether its support count and normalized weight are high or low, yielding four template styles: \emph{well-established}, \emph{common}, \emph{promising}, and \emph{weak}. 
Thus, Auto-Naturalization can distinguish strong, common, emerging, and weak signals while remaining inexpensive, deterministic, and easy to audit. 
The naturalization templates used in our framework are provided in Appendix~\ref{app:naturalization_templates}.

\paragraph{LLM-Naturalization.}
LLM-Naturalization uses an LLM to rewrite structured evidence into a coherent natural-language summary. 
Given the retrieved node-level edges and multi-hop paths, the naturalizer is instructed to group logically related evidence, explain how different pieces of evidence may support or weaken the hypothesis, and optionally add highly relevant background knowledge. 
To reduce unsupported generation, we require the naturalizer to separate graph-grounded evidence from any additional background knowledge it introduces. 
Compared with Auto-Naturalization, LLM-Naturalization is more flexible and can better organize heterogeneous evidence, but it introduces additional inference cost.

\subsection{LLM Hypothesis Review Agent}
\label{sec:review_agent}

The final component of HyGRAIL is an LLM review agent that evaluates each routed hypothesis using the naturalized evidence. 
The review prompt contains the hypothesis, the naturalized evidence, and instructions that encourage the model to assess scientific plausibility based on the provided evidence rather than unsupported assumptions~\citep{wei2022cot,yao2023react,zheng2023llmjudge,zhou2024hypothesisgeneration,qi2024biomedicalhypothesis}. 

For each reviewed hypothesis $h$, the agent outputs a binary decision $d_h \in \{\mathrm{True}, \mathrm{False}\}$ and a confidence score $q_h \in [0,100]$. 
We use both outputs because either one alone can be unreliable. 
A binary decision provides a direct accept/reject judgment, but it does not expose the model's degree of uncertainty. 
A scalar confidence score provides a ranking signal, but LLM confidence estimates can be miscalibrated or concentrated in narrow score ranges~\citep{kadavath2022language,tian2023just}. 
Combining both signals allows HyGRAIL to reject clearly implausible hypotheses while retaining a validation-calibrated threshold over plausible ones.

The confidence threshold $\gamma$ is selected on validation hypotheses routed to the LLM. 
For a test-time routed hypothesis, HyGRAIL accepts it only if $d_h=\mathrm{True}$ and $q_h \geq \gamma$. 
Together with GNN triage, the final prediction rule is: predict $0$ when $s_h<a^\ast$, predict $1$ when $s_h>b^\ast$, and otherwise use the LLM review decision. 
Thus, the GNN handles easy hypotheses at low cost, while the LLM review agent is invoked only for ambiguous hypotheses where natural-language scientific reasoning is expected to add value.
\section{Experimental Setup}
\label{sec:experiments}

\noindent\textbf{Dataset.}
We evaluate HyGRAIL on MatKG~\citep{venugopal2024matkg}, a large-scale materials-science KG automatically constructed from scientific literature.
MatKG is a suitable testbed because it contains typed scientific entities and relations, such as chemicals, properties, and applications, together with edge support counts derived from literature evidence.
To keep the experiments computationally affordable, we sample a 3,000-node subgraph from MatKG while approximately preserving graph density, node type distributions, and edge type distributions.
We use seven MatKG hypothesis types for evaluation; detailed subgraph statistics are provided in Appendix~\ref{app:subkg_stats}.

\noindent\textbf{Data splits.}
For each MatKG edge, we assign a timestamp using the publication time of the oldest paper contributing to its edge count.
For each hypothesis type, observed positive edges are sorted by timestamp and split into train, validation, and test sets with a ratio of 7:1:2.
Within each split, we sample unlinked typed node pairs from the same hypothesis space as negatives, maintaining a positive-to-negative ratio of 1:20 to simulate the highly sparse discovery setting.
Validation and test hypothesis edges are removed from the training graph to prevent label leakage.
The validation set is used for all threshold selection, and the test set is used only for final evaluation.

\noindent\textbf{GNN backbones.}
We instantiate the GNN triage module with three heterogeneous graph backbones:
\begin{itemize}[leftmargin=1.2em, nosep]
    \item \textbf{HeteroConv}: A simple heterogeneous message-passing baseline that applies relation-specific convolutions over typed MatKG edges and aggregates incoming relation-specific representations by mean pooling.
    \item \textbf{HGT}: A stronger heterogeneous encoder that uses type-specific projections and multi-head attention to weight messages from different node and edge types~\citep{hu2020hgt}.
    \item \textbf{R-GCN}: A relational GNN backbone that applies basis-decomposed relation-specific convolutions and scores candidate hypotheses with an MLP over endpoint, relation, and local graph features~\citep{schlichtkrull2018rgcn}.
\end{itemize}
Implementation details and hyperparameters are provided in Appendix~\ref{app:gnn_implementation}.

\noindent\textbf{LLM review models.}
We evaluate four open-weight LLMs with different sizes and families as hypothesis review agents:
Qwen3-4B-Instruct-2507, Qwen3-14B, Ministral-3B-Reasoning, and Ministral-14B-Reasoning~\citep{yang2025qwen3,liu2026ministral3}.
All prompts and naturalization templates are provided in Appendix~\ref{app:prompts_templates}.

\noindent\textbf{Baselines.}
In addition to GNN methods, we compare HyGRAIL with the following variants or prior methods as baselines.
\begin{itemize}[leftmargin=1.2em, nosep]
    \item\textbf{HyGRAIL-no\_evidence(NE).} removes the evidence from HyGRAIL, allowing the LLM to review each routed hypothesis using only the hypothesis text; we use it to evaluate the effectiveness of evidence for hypothesis verification.
    \item \textbf{Topological--Semantic Hybrid(TSH).}
Following \citet{marwitz2026conceptgraph}, this baseline combines local topological descriptors and MatSciBERT entity-name embeddings with validation-tuned weights, omitting the original LLM-based concept extraction stage since our input is already a curated KG.
    \item \textbf{KG-FM.}
Following \citet{bai2025kgfm}, this baseline augments an LLM reviewer with KG-retrieved context and applies it directly to MatKG hypothesis verification.
\end{itemize}

\noindent\textbf{Metrics.}
We report recall and F1 score on the test set, where all thresholds are selected to maximize F1 score on the validation set.
Since HyGRAIL is designed to reduce unnecessary LLM inference, we also report the LLM call rate, defined as the fraction of test hypotheses routed to the LLM review agent.
\begin{table*}[t]
\vspace{-15pt}
\centering
\caption{
\small
\textbf{Main results on MatKG hypothesis verification.}
We report recall and F1 score on the test set.
TSH is independent of both GNN and LLM choices, while KG-FM depends on the LLM reviewer but not the GNN backbone.
For HyGRAIL variants, results are reported across three GNN backbones and four LLM review models.
All highlights are applied to F1 only: bold denotes the overall best result, underline denotes the best result within each GNN block, and italics denote the best result for each LLM reviewer among LLM-conditioned methods.
}
\vspace{-5pt}
\resizebox{0.8\textwidth}{!}{%
\begin{tabular}{l||cc|cc|cc|cc}
        \Xhline{1pt}
        \rowcolor{paleaqua}
        \multicolumn{1}{c||}{\textbf{Method}} 
        & \multicolumn{8}{c}{\textbf{Prior Baselines}} \\
        \hline
        \rowcolor{paleaqua}
        \multicolumn{1}{c||}{\textbf{LLM}} 
        & \multicolumn{2}{c|}{\textbf{Qwen3-4B}}
        & \multicolumn{2}{c|}{\textbf{Qwen3-14B}}
        & \multicolumn{2}{c|}{\textbf{Ministral-3B}}
        & \multicolumn{2}{c}{\textbf{Ministral-14B}} \\
        \rowcolor{paleaqua}
        \textbf{Method}
        & \textbf{Rec.} & \textbf{F1}
        & \textbf{Rec.} & \textbf{F1}
        & \textbf{Rec.} & \textbf{F1}
        & \textbf{Rec.} & \textbf{F1} \\
        \hline \hline
    \textbf{TSH}
    & \multicolumn{8}{c}{Rec. 0.379 \quad F1 0.187} \\
    \textbf{KG-FM}
    & 0.089  & 0.113
    & 0.012 & 0.023
    & 0.245 & 0.149
    & 0.173 & 0.173 \\

        \Xhline{1pt}
        \rowcolor{paleaqua}
        \multicolumn{1}{c||}{\textbf{GNN Backbone}} 
        & \multicolumn{8}{c}{\textbf{HeteroConv}} \\
        \hline
        \rowcolor{paleaqua}
        \multicolumn{1}{c||}{\textbf{LLM}} 
        & \multicolumn{2}{c|}{\textbf{Qwen3-4B}}
        & \multicolumn{2}{c|}{\textbf{Qwen3-14B}}
        & \multicolumn{2}{c|}{\textbf{Ministral-3B}}
        & \multicolumn{2}{c}{\textbf{Ministral-14B}} \\
        \rowcolor{paleaqua}
        \textbf{Method}
        & \textbf{Rec.} & \textbf{F1}
        & \textbf{Rec.} & \textbf{F1}
        & \textbf{Rec.} & \textbf{F1}
        & \textbf{Rec.} & \textbf{F1} \\
        \hline \hline
        \textbf{GNN}
        & \multicolumn{8}{c}{Rec. 0.868 \quad F1 0.093} \\
        \textbf{HyGRAIL-NE}
        & 0.450 & 0.153
        & 0.441 & 0.159
        & 0.481 & 0.144
        & 0.485 & 0.150 \\
        \hline \hline
        \textbf{HyGRAIL-Auto}
        & 0.512 & \underline{0.219}
        & 0.512 & 0.215
        & 0.398 & \textit{0.179}
        & 0.472 & 0.207 \\
        \textbf{HyGRAIL-LLM}
        & 0.479 & 0.206
        & 0.443 & 0.194
        & 0.394 & 0.173
        & 0.450 & 0.197 \\

        \Xhline{1pt}
        \rowcolor{paleaqua}
        \multicolumn{1}{c||}{\textbf{GNN Backbone}} 
        & \multicolumn{8}{c}{\textbf{HGT}} \\
        \hline
        \rowcolor{paleaqua}
        \multicolumn{1}{c||}{\textbf{LLM}} 
        & \multicolumn{2}{c|}{\textbf{Qwen3-4B}}
        & \multicolumn{2}{c|}{\textbf{Qwen3-14B}}
        & \multicolumn{2}{c|}{\textbf{Ministral-3B}}
        & \multicolumn{2}{c}{\textbf{Ministral-14B}} \\
        \rowcolor{paleaqua}
        \textbf{Method}
        & \textbf{Rec.} & \textbf{F1}
        & \textbf{Rec.} & \textbf{F1}
        & \textbf{Rec.} & \textbf{F1}
        & \textbf{Rec.} & \textbf{F1} \\
        \hline \hline
        \textbf{GNN}
        & \multicolumn{8}{c}{Rec. 0.942 \quad F1 0.093} \\
        \textbf{HyGRAIL-NE}
        & 0.568 & 0.103
        & 0.597 & 0.106
        & 0.577 & 0.102
        & 0.588 & 0.103 \\
        \hline \hline
        \textbf{HyGRAIL-Auto}
        & 0.617 & \underline{0.117}
        & 0.611 & 0.115
        & 0.546 & 0.105
        & 0.579 & 0.110 \\
        \textbf{HyGRAIL-LLM}
        & 0.626 & \underline{0.117}
        & 0.557 & 0.106
        & 0.548 & 0.105
        & 0.557 & 0.106 \\

        \Xhline{1pt}
        \rowcolor{paleaqua}
        \multicolumn{1}{c||}{\textbf{GNN Backbone}} 
        & \multicolumn{8}{c}{\textbf{R-GCN}} \\
        \hline
        \rowcolor{paleaqua}
        \multicolumn{1}{c||}{\textbf{LLM}} 
        & \multicolumn{2}{c|}{\textbf{Qwen3-4B}}
        & \multicolumn{2}{c|}{\textbf{Qwen3-14B}}
        & \multicolumn{2}{c|}{\textbf{Ministral-3B}}
        & \multicolumn{2}{c}{\textbf{Ministral-14B}} \\
        \rowcolor{paleaqua}
        \textbf{Method}
        & \textbf{Rec.} & \textbf{F1}
        & \textbf{Rec.} & \textbf{F1}
        & \textbf{Rec.} & \textbf{F1}
        & \textbf{Rec.} & \textbf{F1} \\
        \hline \hline
        \textbf{GNN}
        & \multicolumn{8}{c}{Rec. 0.38 \quad F1 0.107} \\
        \textbf{HyGRAIL-NE}
        & 0.273 & 0.116
        & 0.169 & 0.147
        & 0.312 & 0.122
        & 0.506 & 0.132 \\
        \hline \hline
        \textbf{HyGRAIL-Auto}
        & 0.351 & \textbf{\underline{\textit{0.429}}}
        & 0.351 & \textit{0.362}
        & 0.091 & 0.159
        & 0.286 & \textit{0.333} \\
        \textbf{HyGRAIL-LLM}
        & 0.377 & 0.337
        & 0.234 & 0.319
        & 0.974 & 0.092
        & 0.156 & 0.247 \\

        \Xhline{1pt}
    \end{tabular}
}
\label{tab:main_results}
\vspace{-10pt}
\end{table*}

\section{Results and Analysis}
\label{sec:results}

\subsection{Main Results}

Table~\ref{tab:main_results} reports the main results on MatKG hypothesis verification.
HyGRAIL achieves the best overall performance with R-GCN and Qwen3-4B, reaching an F1 score of $0.429$.
This improves over TSH by $0.242$ F1 points, over the best KG-FM result by $0.256$, and over the corresponding R-GCN-only baseline by $0.322$, showing that graph scores alone are insufficient in this sparse setting.

Retrieved evidence is consistently beneficial over the evidence-free variant.
Under HeteroConv, HyGRAIL-Auto improves over HyGRAIL-NE across all four LLM reviewers, with F1 gains of $0.035$--$0.066$.
With R-GCN, the gains are larger: $0.313$ and $0.215$ F1 points for Qwen3-4B and Qwen3-14B.
These results confirm that hypothesis-guided graph evidence and evidence naturalization are central to HyGRAIL's effectiveness.

HyGRAIL-Auto is generally stronger than HyGRAIL-LLM, suggesting that deterministic template-based naturalization provides more concise and controlled evidence descriptions.
Overall, the results support our design of using GNNs for cost-aware triage and naturalized graph evidence for LLM review on ambiguous hypotheses.

\subsection{LLM Call Reduction}

Table~\ref{tab:llm_call_rate} reports the fraction of test hypotheses routed to the LLM review agent.
By filtering confident low-score and high-score hypotheses before LLM review, HyGRAIL substantially reduces the number of LLM calls across all GNN backbones.
On average, GNN triage lowers the LLM call rate to \textbf{45.64\%}.
This confirms that GNN triage is not only a predictive component, but also an effective cost-control mechanism for large-scale scientific hypothesis discovery.

\begin{table}[t]
\centering
\small
\caption{
\textbf{GNN triage reduces LLM review cost.}
LLM call rate is the fraction of test hypotheses routed to the LLM review agent.
}
\vspace{-5pt}
\begin{tabular}{lcc}
\toprule
\textbf{GNN Backbone} & \textbf{LLM Call Rate} & \textbf{Reduction} \\
\midrule
HeteroConv & 53.45\% & 46.55\% \\
HGT & 29.99\% & 70.01\% \\
R-GCN & 53.49\% & 46.51\% \\
\bottomrule
\end{tabular}
\label{tab:llm_call_rate}
\vspace{-15pt}
\end{table}

\subsection{Ablation Study}
\label{sec:ablation}

We conduct ablations with HGT and Qwen3-4B / Qwen3-14B to examine how retrieved graph evidence affects hypothesis verification.

\begin{figure}[t]
    \centering
    \includegraphics[width=\linewidth]{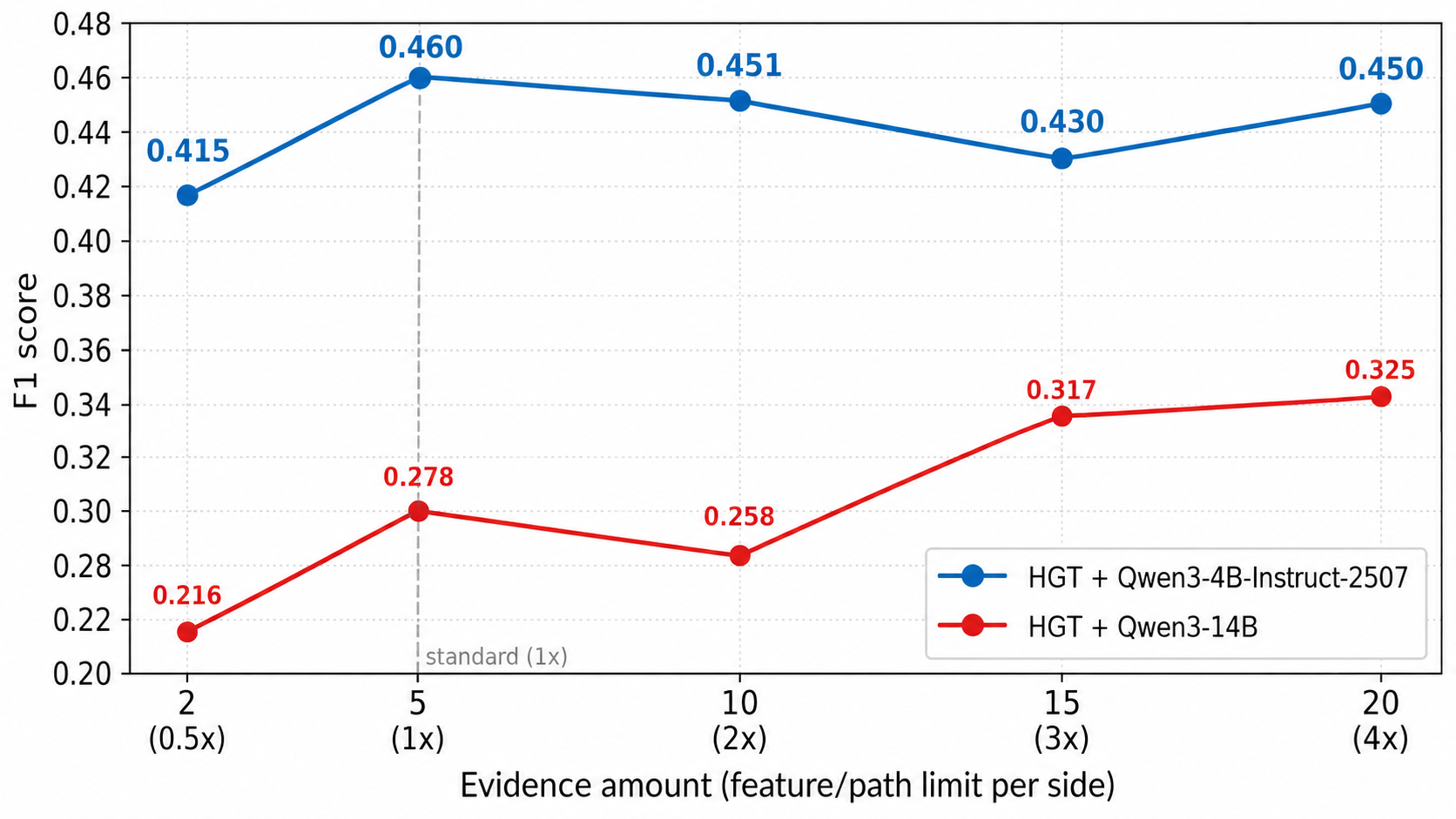}
    \vspace{-5pt}
    \caption{
    \textbf{Evidence is useful, but more evidence is not always better.}
The default budget improves over smaller evidence, while excessive evidence yields inconsistent gains and can hurt stronger reviewers.
    }
    \label{fig:ablation_evidence_quantity}
    \vspace{-10pt}
\end{figure}

\paragraph{Effect of evidence quantity.}
Figure~\ref{fig:ablation_evidence_quantity} shows that retrieved evidence is essential, but simply increasing the amount of evidence is not always beneficial.
Compared with the smaller evidence budget, the default HyGRAIL setting improves F1 from $0.415$ to $0.460$ for Qwen3-4B and from $0.216$ to $0.278$ for Qwen3-14B.
However, further increasing the evidence budget yields unstable gains: Qwen3-4B drops below its default-budget performance under all larger budgets, while Qwen3-14B benefits only at larger budgets after an initial drop at $2\times$ evidence.
This suggests that excessive retrieved context may introduce noise rather than useful scientific signal, supporting our design choice of using a compact, hypothesis-guided evidence set instead of maximizing retrieval quantity.

\begin{figure}[t]
    \centering
    \includegraphics[width=\linewidth]{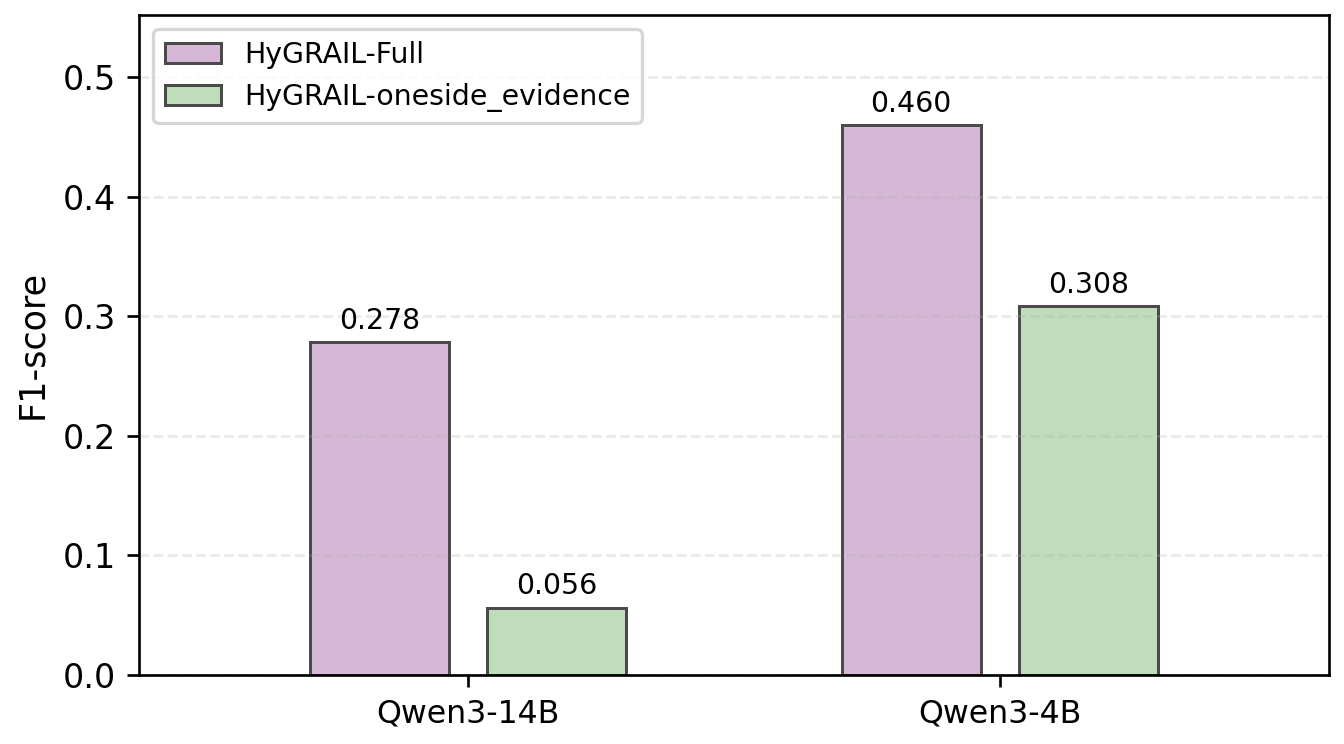}
    \vspace{-5pt}
    \caption{
    \textbf{Evidence from both endpoints is crucial.}
    Full HyGRAIL substantially outperforms the one-sided evidence variant under both LLM reviewers.
    }
    \label{fig:ablation_oneside}
    \vspace{-15pt}
\end{figure}

\paragraph{Effect of two-sided endpoint evidence.}
Figure~\ref{fig:ablation_oneside} compares HyGRAIL with a one-sided variant that keeps the total amount of node-level evidence unchanged but retrieves evidence from only one endpoint.
Full HyGRAIL substantially outperforms the one-sided variant for both LLM reviewers, improving F1 by \textbf{0.222} for Qwen3-14B and \textbf{0.152} for Qwen3-4B.
This indicates that the two endpoints of a candidate link provide complementary scientific context, and that balanced evidence from both sides is important for reliable hypothesis verification.

\section{Related Work}
\label{sec:related}

\noindent\textbf{Scientific KGs and hypothesis discovery.}
Computational scientific discovery has long studied how hidden hypotheses can be inferred by connecting fragmented literature evidence~\citep{swanson1986undiscovered}. 
Scientific KGs scale this process by structuring entities and relations from literature or curated databases, including biomedical KGs~\citep{kilicoglu2012semmeddb,himmelstein2017hetionet}, scholarly KGs~\citep{jaradeh2019orkg}, and materials-science KGs~\citep{venugopal2024matkg}. 
These resources support graph- and literature-based hypothesis generation~\citep{sosa2020literature,sybrandt2020agatha}, while recent work also explores LLMs for scientific hypothesis generation~\citep{zhou2024hypothesisgeneration,qi2024biomedicalhypothesis,yang2024moosechem}. 
HyGRAIL differs by focusing on evidence-grounded verification of typed missing links rather than open-ended hypothesis generation.

\noindent\textbf{Graph-based link prediction.}
KG completion methods, including embedding models~\citep{bordes2013transe,yang2015distmult,trouillon2016complex}, relational and heterogeneous GNNs~\citep{schlichtkrull2018rgcn,wang2019han,hu2020hgt}, and path-based reasoning methods~\citep{lao2010relational,neelakantan2015compositional,das2018minerva}, provide efficient tools for missing-relation prediction. 
However, they mainly rely on observed topology and relation statistics, which can be insufficient for sparse and ambiguous scientific hypotheses. 
HyGRAIL instead uses GNNs as a cost-aware triage module and invokes LLM-based evidence reasoning only for graph-ambiguous cases.

\noindent\textbf{Retrieval-augmented LLM reasoning.}
Retrieval-augmented language models ground generation in external evidence~\citep{lewis2020rag,guu2020realm,karpukhin2020dpr}, and KG-guided RAG further uses graph relations to organize retrieved information~\citep{zhu2025kg2rag}. 
KG-to-text and data-to-text methods verbalize structured triples into natural language~\citep{gardent2017webnlg,ribeiro2020graph2text}, complementing LLM reasoning and agentic prompting~\citep{wei2022cot,yao2023react}. 
Unlike prior work on QA, text generation, or free-form reasoning, HyGRAIL retrieves and naturalizes typed graph evidence for cost-aware verification of specific scientific hypotheses.
\section{Conclusion}
\label{sec:conclusion}

We introduced \textbf{HyGRAIL}, a hybrid framework for cost-aware and evidence-grounded scientific hypothesis discovery over incomplete heterogeneous KGs. 
By combining GNN-based triage, hypothesis-guided graph evidence retrieval, evidence naturalization, and LLM-based review, HyGRAIL improves hypothesis verification while reducing unnecessary LLM calls. 
Experiments on MatKG and ablations on evidence quantity and sidedness show that retrieved graph evidence is crucial for reliable verification. 
HyGRAIL further supports evidence-grounded hypothesis prioritization for expert inspection or experimental validation. 
Overall, our results highlight the complementary value of scalable graph learning and evidence-grounded LLM reasoning for scientific discovery.

\section{Limitations}
\label{sec:limitations}

Our study has several limitations. 
First, our evaluation follows a closed-world link prediction protocol, where held-out observed edges are treated as positives and sampled unlinked pairs are treated as negatives. 
This protocol is standard for KG link prediction, but an unobserved edge does not necessarily imply scientific invalidity. 
Therefore, hypotheses accepted by HyGRAIL should be interpreted as candidates for expert inspection or experimental validation rather than definitive scientific facts.

Second, HyGRAIL relies on predefined hypothesis types and evidence edge type sets for graph evidence retrieval. 
This design improves controllability and interpretability, but transferring the framework to a new KG schema may require lightweight domain-specific configuration. 
Future work could explore more automated evidence type selection and schema adaptation.

Third, the LLM review stage may be affected by model choice, prompting details, and confidence calibration. 
Although HyGRAIL mitigates this issue by using validation-selected thresholds and explicit evidence grounding, further calibration or ensemble-based review could make the final decisions more robust.

\section{Ethical Considerations}
\label{sec:ethics}

This work studies scientific hypothesis verification over a publicly available scientific knowledge graph and does not involve human subjects, personal data, private user information, or demographic attributes. 
The proposed framework is intended to assist scientific hypothesis prioritization rather than replace expert judgment or experimental validation. 
Because hypotheses accepted by HyGRAIL are model-generated candidates, they should be interpreted as suggestions for further expert review. 
We do not foresee direct ethical risks beyond standard concerns associated with automated scientific decision support, such as over-reliance on model outputs without domain expert validation. We use publicly available scientific artifacts and open-weight models only for research evaluation, cite their original creators, and do not redistribute the original artifacts.
\bibliography{custom}

\appendix

\tcbset{
  promptstyle/.style={
    colback=brown!10,
    colframe=brown!50!black,
    fonttitle=\bfseries,
    coltitle=white,
    colbacktitle=brown!50!black,
    boxrule=0.75mm,
    coltext=black,
    width=\textwidth,
    fontupper=\small,
  }
}

\section{Details of MatKG and Data Construction}
\label{app:matkg_details}

\subsection{Statistics of the Sampled MatKG Subgraph}
\label{app:subkg_stats}

We sample a 3,000-node subgraph from MatKG while approximately preserving graph density, node type distribution, and edge type distribution. 
Detailed statistics of the sampled subgraph are shown in Tables~\ref{tab:app_node_stats} and~\ref{tab:app_edge_stats}. 
The graph density of the sampled subgraph is $1.097 \times 10^{-3}$.

\begin{table}[h]
\centering
\small
\begin{tabular}{lrr}
\toprule
\textbf{Node Type} & \textbf{\# Nodes} & \textbf{Ratio} \\
\midrule
\texttt{CHM} & 661 & 22.03\% \\
\texttt{PRO} & 1,013 & 33.77\% \\
\texttt{APL} & 419 & 13.97\% \\
\texttt{SMT} & 221 & 7.37\% \\
\texttt{DSC} & 205 & 6.83\% \\
\texttt{CMT} & 448 & 14.93\% \\
\texttt{SPL} & 33 & 1.10\% \\
\midrule
\textbf{Total} & 3,000 & 100\% \\
\bottomrule
\end{tabular}
\caption{
Node type statistics of the sampled MatKG subgraph.
}
\label{tab:app_node_stats}
\end{table}

\begin{table}[h]
\centering
\small
\begin{tabular}{lrr}
\toprule
\textbf{Edge Type} & \textbf{\# Edges} & \textbf{Ratio} \\
\midrule
\texttt{CHM--APL} & 194 & 1.96\% \\
\texttt{CHM--PRO} & 621 & 6.29\% \\
\texttt{PRO--APL} & 216 & 2.19\% \\
\texttt{SMT--PRO} & 138 & 1.40\% \\
\texttt{CHM--SMT} & 152 & 1.54\% \\
\texttt{CHM--DSC} & 304 & 3.08\% \\
\texttt{CHM--CMT} & 385 & 3.90\% \\
Other edge types & 7,864 & 79.65\% \\
\midrule
\textbf{Total} & \ 9,874 & 100\% \\
\bottomrule
\end{tabular}
\caption{
Edge type statistics of the sampled MatKG subgraph.
}
\label{tab:app_edge_stats}
\end{table}

\subsection{Data Split Statistics}
\label{app:data_split_stats}

For each hypothesis type, observed positive edges are split into training, validation, and test sets with a ratio of 7:1:2 according to edge timestamps. 
Within each split, negative examples are sampled from unlinked typed node pairs with a positive-to-negative ratio of 1:20. 
Table~\ref{tab:app_split_stats} summarizes the number of cases for each hypothesis type and split.

\begin{table*}[t]
\centering
\small
\begin{tabular}{lrrrrrrrrr}
\toprule
\multirow{2}{*}{\textbf{Hypothesis Type}} 
& \multicolumn{3}{c}{\textbf{Train}} 
& \multicolumn{3}{c}{\textbf{Validation}} 
& \multicolumn{3}{c}{\textbf{Test}} \\
\cmidrule(lr){2-4}
\cmidrule(lr){5-7}
\cmidrule(lr){8-10}
& \textbf{Pos.} & \textbf{Neg.} & \textbf{Total}
& \textbf{Pos.} & \textbf{Neg.} & \textbf{Total}
& \textbf{Pos.} & \textbf{Neg.} & \textbf{Total} \\
\midrule
\texttt{CHM--APL} & 137 & 2,714 & 2,851 & 18 & 390 & 408 & 39 & 776 & 815 \\
\texttt{CHM--PRO} & 437 & 8,691 & 9,128 & 58 & 1,246 & 1,304 & 126 & 2,483 & 2,609 \\
\texttt{PRO--APL} & 152 & 3,023 & 3,175 & 21 & 432 & 453 & 43 & 865 & 908 \\
\texttt{SMT--PRO} & 98 & 1,930 & 2,028 & 13 & 277 & 290 & 27 & 553 & 580 \\
\texttt{CHM--SMT} & 106 & 2,128 & 2,234 & 15 & 304 & 319 & 31 & 608 & 639 \\
\texttt{CHM--DSC} & 213 & 4,255 & 4,468 & 31 & 608 & 639 & 60 & 1,217 & 1,277 \\
\texttt{CHM--CMT} & 270 & 5,389 & 5,659 & 38 & 771 & 809 & 77 & 1,540 & 1,617 \\
\midrule
Total & 1,413 & 28,130 & 29,543 & 194 & 4,028 & 4,222 & 403 & 8,042 & 8,445 \\
\bottomrule
\end{tabular}
\caption{
Data split statistics for the seven MatKG hypothesis types.
Each split maintains a positive-to-negative ratio of 1:20.
}
\label{tab:app_split_stats}
\end{table*}

\subsection{Evidence Edge Types for Node-level Retrieval}
\label{app:evidence_edge_types}

For node-level evidence retrieval, each node type is associated with a predefined set of evidence edge types.
Table~\ref{tab:app_evidence_edge_types} lists the evidence edge type set used for each MatKG node type.

\begin{table*}[t]
\centering
\small
\begin{tabularx}{\textwidth}{lX}
\toprule
\textbf{Node Type} & \textbf{Evidence Edge Types} \\
\midrule
\texttt{CHM} & CHM-PRO, CHM-APL, CHM-SPL, CHM-SMT, CHM-CMT, CHM-DSC, CHM-CHM \\
\texttt{PRO} & PRO-CHM, PRO-SPL, PRO-APL, PRO-SMT, PRO-CMT \\
\texttt{APL} & APL-PRO, APL-CHM \\
\texttt{SMT} & SMT-CHM, SMT-SPL, SMT-PRO, SMT-CMT \\
\texttt{DSC} & DSC-CHM \\
\texttt{CMT} & CMT-PRO, CMT-SPL, CMT-CHM \\
\texttt{SPL} & SPL-CHM, SPL-PRO, SPL-SMT, SPL-CMT \\
\bottomrule
\end{tabularx}
\caption{
Evidence edge type set for each MatKG node type.
}
\label{tab:app_evidence_edge_types}
\end{table*}

\section{Implementation Details}
\label{app:implementation}

\subsection{GNN Implementation Details and Hyperparameter Selection}
\label{app:gnn_implementation}

All GNN backbones are trained separately for each target hypothesis type. 
For each relation, we build the message-passing graph from the sampled MatKG subgraph and remove validation and test positive edges of the target relation from the training graph to avoid label leakage. 
Edge support counts are transformed with $\log(1+c)$ and min-max normalized within each edge relation when the backbone uses edge weights. 
The candidate classifier is trained with binary cross-entropy over the timestamp-based training split with the same 1:20 positive-to-negative ratio described in Section~\ref{sec:experiments}. 
All experiments use Adam with learning rate $10^{-3}$ and random seed 42. 
For evaluation, each GNN produces a score $s_h \in [0,1]$ for a candidate hypothesis $h$; the binary decision threshold is selected on the validation split by maximizing F1 and is then fixed for test evaluation.

\paragraph{HeteroConv.}
The HeteroConv backbone uses learnable node embeddings for each MatKG node type and two PyG \texttt{HeteroConv} layers. 
Each layer contains a separate \texttt{GraphConv} operator for every observed typed edge relation, uses mean aggregation across relations, and applies layer normalization; the first layer is followed by ReLU. 
The hidden dimension is 64, training runs for 100 epochs, and validation metrics are checked every 10 epochs. 
The link decoder computes the dot product between $\ell_2$-normalized endpoint embeddings.

\paragraph{HGT.}
The HGT backbone also starts from learnable type-specific node embeddings, followed by type-specific linear projections and two \texttt{HGTConv} layers. 
We use hidden dimension 64 and 4 attention heads, with layer normalization and ReLU after each HGT layer. 
Training runs for 100 epochs with validation checked every 10 epochs, and the same normalized dot-product decoder is used for candidate-edge scoring.

\paragraph{R-GCN.}
For the R-GCN-style backbone, all nodes are mapped into one global node-id space and each KG edge is assigned a relation-type id. 
The encoder uses two \texttt{RGCNConv} layers with hidden dimension 96 and 8 basis functions, followed by layer normalization. 
Unlike the other two backbones, this model uses an edge scorer designed for sparse scientific hypothesis links: for each candidate, it concatenates the normalized source embedding, normalized target embedding, elementwise product, absolute embedding difference, a learned target-relation embedding, and six standardized local features. 
These local features are the log weighted degree of each endpoint, a weighted two-hop/shared-neighbor signal, log common-neighbor count, Jaccard similarity, and an Adamic-Adar-style score. 
The concatenated vector is passed through an MLP with two hidden layers and dropout rates 0.15 and 0.10. 
We train for 120 epochs, check validation metrics every 20 epochs, and use Adam with weight decay $10^{-5}$. 
The best R-GCN checkpoint is selected by validation max-F1 before selecting the final validation F1 threshold.

\subsection{LLM Inference Hyperparameter Selection}
\label{app:llm_inference}

\textbf{Serving Stack.} Each of the four models is launched as an independent vLLM instance exposing an OpenAI-compatible \texttt{/v1/chat/completions} endpoint on a dedicated GPU and port, with no tensor parallelism. 

\textbf{Per-Model Serving and Token Budgets.} vLLM launch settings and client-side decoding budgets are summarised in Table~\ref{tab:vllm_serving}.

\begin{table}[h]
\centering
\small
\setlength{\tabcolsep}{3pt}
\caption{Per-model vLLM serving and decoding settings. }
\label{tab:vllm_serving}
\resizebox{\columnwidth}{!}{%
\begin{tabular}{lcccc}
\toprule
\textbf{Model} & \textbf{gpu-mem} & \textbf{max-len} & \textbf{max-seq} & \textbf{new-tok} \\
\midrule
Qwen3-4B-Instruct-2507  & 0.90 & 32{,}768 & 64 & 1024 \\
Qwen3-14B               & 0.90 & 32{,}768 & 32 & 1024 \\
Mistral-3B-reasoning    & 0.85 & default  & 64 & 4096 \\
Magistral-Small-14B     & 0.90 & 32{,}768 & 16 & 4096 \\
\bottomrule
\end{tabular}
}
\end{table}

\textbf{Decoding.} Sampling parameters are shared across all models:
\begin{itemize}
\itemsep0em
\item \texttt{temperature} = 0.2
\item \texttt{top\_p} = 0.9
\item \texttt{request\_timeout} = 180\,s
\end{itemize}

\textbf{Thinking Mode Control.} Reasoning behaviour is governed at the client side through \texttt{chat\_template\_kwargs=\{"enable\_thinking":...\}}, rather than through the server flag \texttt{--reasoning-parser}.
\begin{itemize}
\itemsep0em
\item \textbf{Qwen3-14B}: thinking explicitly disabled.
\item \textbf{Qwen3-4B-Instruct-2507}: no thinking mode by construction.
\item \textbf{Mistral / Magistral reasoning models}: native thinking chain retained, parameter left unset.
\end{itemize}

\textbf{Output Contract.} Every model is required to emit a single-line JSON object of the form \texttt{\{"reasoning":..., "review":"yes/no", "confidence":0.XX\}}. The semantic convention is \texttt{review="yes"} iff \texttt{confidence} $\geq 0.5$, where \texttt{confidence} denotes $P(\text{hypothesis is true})$.

\textbf{Inference Matrix.} The full sweep covers 7 relations $\times$ 4 models $\times$ 3 evidence modes (\textit{no-evidence}, \textit{template-evidence}, \textit{llm-evidence}). The \textit{no-evidence} mode uses a dedicated prompt that elicits the model prior, while \textit{template-evidence} and \textit{llm-evidence} share a prompt that conditions the model on the KG-derived evidence vocabulary.

\section{The Use of Large Language Models}
During the writing of this paper, we used the GPT-5 Mini model for text polishing and grammatical corrections to enhance the readability of the manuscript.

\section{Prompts and Naturalization Templates}
\label{app:prompts_templates}
\subsection{LLM Prompts}
\label{app:llm_prompts}

We provide the prompts used for evidence naturalization and hypothesis review in Figures~\ref{fig:app_naturalization_prompt} and~\ref{fig:app_review_prompt}, respectively.

\subsection{Naturalization Templates}
\label{app:naturalization_templates}

Auto-Naturalization uses deterministic templates to convert structured graph evidence into natural-language statements.
Table~\ref{tab:app_naturalization_templates} summarizes the template styles used in our framework.

\begin{figure*}[t]
\begin{center}
\begin{tcolorbox}[promptstyle, title=Evidence Naturalization Prompt]

\textbf{[System Prompt]}

You are a scientific evidence writer for materials-science KG outputs.

\medskip
\noindent\textbf{Task.}
Convert structured KG evidence into a coherent natural-language summary for downstream hypothesis validation.

\medskip
\noindent\textbf{Rules.}
\begin{enumerate}[leftmargin=1.2em, nosep]
    \item First summarize graph-grounded evidence only, grouping logically related evidence into coherent themes.
    \item Explain how each evidence group supports or weakens the hypothesis; if signals are mixed, say so explicitly.
    \item You may add highly relevant background knowledge only when it materially helps interpret the graph evidence.
    \item Any background knowledge must be kept separate from graph-grounded evidence.
    \item Faithfully preserve provided entities, paths, and numerical values, including \texttt{norm}, \texttt{count}, and \texttt{score}.
    \item Use neutral, descriptive wording. Do not make a yes/no verdict and do not predict plausibility.
    \item Do not invent citations, experiments, or unsupported mechanisms.
    \item If evidence is weak, sparse, has missing paths, or only goes through generic hubs such as \textit{Structure}, \textit{Film}, \textit{Layer}, \textit{Interface}, or \textit{Device}, state that explicitly.
    \item Use the exact vocabulary \texttt{norm}, \texttt{count}, and \texttt{cat} when describing features so downstream judging stays calibrated.
\end{enumerate}

\medskip
\noindent\textbf{Output.}
Return plain prose only: no markdown, no bullets, and no JSON. 
If outside knowledge is added, keep it in a clearly separate paragraph beginning with ``Background knowledge (optional):''.

\vspace{0.5em}
\hrule
\vspace{0.5em}

\textbf{[User Prompt Template]}

Convert structured knowledge graph evidence into a coherent natural-language summary for hypothesis validation.

\medskip
\noindent\textbf{Hypothesis.}
\begin{itemize}[leftmargin=1.2em, nosep]
    \item Left entity: \texttt{\{left\_entity\}} (\texttt{\{left\_type\_code\}}: \texttt{\{left\_type\_label\}})
    \item Right entity: \texttt{\{right\_entity\}} (\texttt{\{right\_type\_code\}}: \texttt{\{right\_type\_label\}})
    \item Hypothesis type: \texttt{\{hypothesis\_type\}}
    \item Hypothesis statement: \texttt{\{hypothesis\_statement\}}
\end{itemize}

\medskip
\noindent\textbf{Evidence from Knowledge Graph.}

\texttt{\{candidate\_signal\}}

\medskip
\noindent Raw structured evidence:

\texttt{\{evidence\_text\}}

\medskip
\noindent\textbf{Feature Interpretation Context.}
Each feature provides \texttt{norm}, \texttt{count}, and \texttt{cat}:
\begin{itemize}[leftmargin=1.2em, nosep]
    \item \texttt{norm}: local normalized association strength for that source node and relation type.
    \item \texttt{count}: raw co-occurrence support from the corpus.
    \item \texttt{cat}: edge relevance category, where I indicates strongest relevance, II medium relevance, and III lower relevance or noisier evidence.
\end{itemize}

Use the following four-way interpretation:
\begin{enumerate}[leftmargin=1.2em, nosep]
    \item High \texttt{norm} + high/moderate \texttt{count}: well-established and reliable evidence.
    \item High \texttt{norm} + low \texttt{count}: promising but under-explored evidence.
    \item Low/moderate \texttt{norm} + high/moderate \texttt{count}: common but less discriminative evidence.
    \item Low \texttt{norm} + low \texttt{count}: weak evidence.
\end{enumerate}

Thresholds:
\begin{itemize}[leftmargin=1.2em, nosep]
    \item norm high $\geq$ \texttt{\{norm\_high:.4f\}}
    \item norm moderate $\geq$ \texttt{\{norm\_med:.4f\}}
    \item count high $\geq$ \texttt{\{count\_high:.1f\}}
    \item count moderate $\geq$ \texttt{\{count\_med:.1f\}}
\end{itemize}

When \texttt{cat=III} or intermediate nodes are generic hubs such as \textit{Structure}, \textit{Film}, \textit{Layer}, \textit{Interface}, or \textit{Device}, explicitly describe them as weaker and less discriminative evidence.

\medskip
\noindent\textbf{Conversion Goals.}
\begin{enumerate}[leftmargin=1.2em, nosep]
    \item Group logically related evidence together instead of translating each edge independently.
    \item Explain how each evidence group supports or weakens the hypothesis; if the signals are mixed, say so directly.
    \item Use both node-level edges and multi-hop paths when available, and explain how they relate to each other.
    \item Keep graph-grounded evidence separate from any extra background knowledge.
    \item Add optional background knowledge only if it is highly relevant and materially helps interpret the graph evidence.
    \item Preserve numerical values and the exact vocabulary \texttt{norm}, \texttt{count}, and \texttt{cat}; do not add unsupported scientific claims.
\end{enumerate}

\medskip
\noindent\textbf{Output Format.}
Write 2--4 short paragraphs of plain prose only.
The first paragraph must begin with ``Graph-grounded evidence:'' and summarize grouped evidence from the graph.
Include a paragraph that explains how the evidence supports or weakens the hypothesis, optionally beginning with ``How the evidence relates to the hypothesis:''.
If there is no connecting path, mention that fact plainly without overstating its negative impact.
If extra scientific context is added, put it in a separate paragraph beginning with ``Background knowledge (optional):''.
Do not output markdown, bullet lists, or JSON.

\end{tcolorbox}
\end{center}
\caption{
Prompt used for LLM-based evidence naturalization.
}
\label{fig:app_naturalization_prompt}
\end{figure*}

\begin{figure*}[t]
\begin{center}
\begin{tcolorbox}[promptstyle, title=Hypothesis Review Prompt]

\textbf{[System Prompt]}

You are a scientific hypothesis validator for materials science knowledge graphs.

\medskip
\noindent\textbf{Task.}
You will receive evidence extracted from MatKG about a proposed relationship between two entities.
Your job is to determine whether this relationship is scientifically plausible by grounding your reasoning in:
\begin{enumerate}[leftmargin=1.2em, nosep]
    \item the provided KG evidence, and
    \item your internalized materials-science knowledge for interpretation, synthesis, and calibration.
\end{enumerate}

\medskip
\noindent\textbf{Hypothesis Types.}
\begin{itemize}[leftmargin=1.2em, nosep]
    \item The hypothesis type is given as a typed pair such as \texttt{CHM-APL}, \texttt{CHM-PRO}, \texttt{PRO-APL}, \texttt{SMT-PRO}, \texttt{CHM-CHM}, \texttt{APL-PRO}, \texttt{CHM-SMT}, \texttt{CHM-DSC}, \texttt{CHM-CMT}, or another valid type pair.
    \item Entity type codes include \texttt{CHM} for material/chemical, \texttt{APL} for application, \texttt{PRO} for property, \texttt{SMT} for synthesis method, \texttt{CMT} for characterization method, \texttt{DSC} for descriptor, and \texttt{SPL} for sample.
    \item Treat the hypothesis as a general scientific relation between the left entity and the right entity under the provided type pair.
\end{itemize}

\medskip
\noindent\textbf{Evidence Vocabulary.}
Use the following exact terms in reasoning:
\begin{itemize}[leftmargin=1.2em, nosep]
    \item \texttt{norm}: local normalized association strength of the feature edge.
    \item \texttt{count}: raw co-occurrence support from the corpus.
    \item \texttt{cat}: edge relevance category, where I is strongest, II is medium, and III is weakest or noisier.
\end{itemize}

Feature quality is interpreted by the following four-way rule:
\begin{enumerate}[leftmargin=1.2em, nosep]
    \item high \texttt{norm} + high/moderate \texttt{count}: well-established;
    \item high \texttt{norm} + low \texttt{count}: promising but under-explored;
    \item low/moderate \texttt{norm} + high/moderate \texttt{count}: common but less discriminative;
    \item low \texttt{norm} + low \texttt{count}: weak.
\end{enumerate}

Connecting paths come with a path score; paths through generic hubs such as \textit{Structure}, \textit{Film}, \textit{Layer}, \textit{Interface}, or \textit{Device} are down-weighted.

\medskip
\noindent\textbf{Reasoning Rules.}
\begin{enumerate}[leftmargin=1.2em, nosep]
    \item Ground your judgment in the KG evidence. Reference specific features, paths, or warnings rather than restating the hypothesis.
    \item Use scientific knowledge to interpret the retrieved evidence, connect related signals, and calibrate uncertainty. Do not replace the evidence with unsupported priors.
    \item Treat \texttt{cat=III} edges and paths through generic hubs as weaker context, but do not map path strength to confidence by itself.
    \item Respect explicit warnings and inconsistencies in the evidence.
    \item Do not fabricate citations, papers, or experiments. Do not invoke external sources.
\end{enumerate}

\medskip
\noindent\textbf{Confidence Semantics.}
\begin{itemize}[leftmargin=1.2em, nosep]
    \item \texttt{confidence} is the probability that the hypothesis is true: $P(\mathrm{review}=\mathrm{yes})$.
    \item \texttt{review} must equal \texttt{yes} iff \texttt{confidence} $\geq 0.5$; otherwise it must equal \texttt{no}.
    \item Do not report high confidence for a \texttt{no} verdict.
\end{itemize}

\medskip
\noindent\textbf{Output Format.}
Return exactly one JSON object on a single line:

\begin{quote}
\ttfamily
\{"reasoning":"<2-3 sentence justification citing specific evidence terms (norm/count/cat/path)>","review":"yes/no","confidence":0.XX\}
\end{quote}

\vspace{0.5em}
\hrule
\vspace{0.5em}

\textbf{[User Prompt Template]}

\medskip
\noindent\textbf{Hypothesis:}

\texttt{\{hypothesis\}}

\medskip
\noindent\textbf{Hypothesis Type:}

\texttt{\{hypothesis\_type\}}

\medskip
\noindent\textbf{Evidence:}

\texttt{\{evidence\}}

\medskip
\noindent\textbf{Task.}
Estimate $P(\text{hypothesis is true})$ given the KG evidence above.
Report it as \texttt{confidence} and set \texttt{review=yes} iff \texttt{confidence} $\geq 0.5$.
When unsure, stay near $0.5$ and do not push confidence in either direction without evidence.
Output one JSON object only.

\end{tcolorbox}
\end{center}
\caption{
Prompt used for LLM hypothesis review.
}
\label{fig:app_review_prompt}
\end{figure*}

\begin{table*}[thbp]
\vspace{-30pt}
\centering
\caption{Summary of Hypothesis and Evidence Naturalization Templates}
\label{tab:app_naturalization_templates}
\renewcommand{\arraystretch}{1.3}
\small
\begin{tabular}{@{}p{0.18\linewidth} p{0.78\linewidth}@{}}
\toprule
\textbf{Category} & \textbf{Template Structure} \\
\midrule

\multicolumn{2}{@{}l}{\textbf{1. Overall Evidence Format}} \\
\midrule
Full Structure & Hypothesis: \{hypothesis\_statement\} \newline
Evidence: The following evidence about \{left\_entity\} and \{right\_entity\} has been extracted from the knowledge graph: \newline
\{left\_entity\_feature\_sentences\} \newline
\{right\_entity\_feature\_sentences\} \newline
\{connecting\_path\_sentences or no-path sentence\} \newline
\{optional\_convergence\_sentence\} \newline
\{optional\_contradiction\_sentence\} \newline
(GNN prediction score: \{score:.4f\}) \\

\midrule
\multicolumn{2}{@{}l}{\textbf{2. Hypothesis Statements}} \\
\midrule
CHM-APL & The material \{left\} can be effectively used for \{right\}. \\
CHM-PRO & The material \{left\} exhibits the property \{right\}. \\
PRO-APL & The property \{left\} is relevant and beneficial to the application \{right\}. \\
SMT-PRO & The synthesis method \{left\} can produce or enhance the property \{right\}. \\
CHM-SMT & The material \{left\} can be synthesized using the method \{right\}. \\
CHM-DSC & The material \{left\} can be described by the descriptor \{right\}. \\
CHM-CMT & The material \{left\} can be characterized using the method \{right\}. \\

\midrule
\multicolumn{2}{@{}l}{\textbf{3. Left-Entity Features}} \\
\midrule
Well-established & \{left\} is strongly characterized by \{right\} (norm=\{norm:.2f\}, count=\{count:.0f\}), representing a well-established and defining trait \\
Promising & \{left\} shows a notable association with \{right\} (norm=\{norm:.2f\}, count=\{count:.0f\}), suggesting a promising but less-explored characteristic \\
Common & \{left\} has some association with \{right\} (norm=\{norm:.2f\}, count=\{count:.0f\}), though this is a common trait shared by many materials \\
Weak & \{left\} has a weak association with \{right\} (norm=\{norm:.2f\}, count=\{count:.0f\}), providing limited discriminative evidence \\

\midrule
\multicolumn{2}{@{}l}{\textbf{4. Right-Entity Features}} \\
\midrule
Well-established & \{left\} fundamentally requires \{right\} (norm=\{norm:.2f\}, count=\{count:.0f\}), a well-documented core requirement \\
Promising & \{left\} appears to benefit from \{right\} (norm=\{norm:.2f\}, count=\{count:.0f\}), an emerging but less-explored requirement \\
Common & \{left\} has some relevance to \{right\} (norm=\{norm:.2f\}, count=\{count:.0f\}), though this is a general association \\
Weak & \{left\} shows weak relevance to \{right\} (norm=\{norm:.2f\}, count=\{count:.0f\}), suggesting peripheral importance \\

\midrule
\multicolumn{2}{@{}l}{\textbf{5. Connecting Paths}} \\
\midrule
Strong & A strong mechanistic link exists: \{material\} possesses \{intermediate\}, which is directly relevant to \{application\} (path score=\{score:.3f\}, indicating robust evidence) \\
Moderate & A plausible connection exists: \{material\} exhibits \{intermediate\}, which relates to \{application\} (path score=\{score:.3f\}, suggesting moderate evidence) \\
Weak & A weak connection exists: \{material\} is linked to \{application\} via \{intermediate\} (path score=\{score:.3f\}), but this provides limited mechanistic support \\
Hub Node & A path exists via \{intermediate\}, but this is a generic/common node (path score=\{score:.3f\}), providing weak discriminative evidence \\

\midrule
\multicolumn{2}{@{}l}{\textbf{6. Structural Context Sentences}} \\
\midrule
No-path & No connecting paths were found in the knowledge graph between \{left\_entity\} and \{right\_entity\}, limiting the mechanistic evidence available. \\
Convergence & Notably, multiple connecting paths converge on similar mechanisms (\{mechanisms\}), strengthening the evidence for this relationship. \\
Contradiction & However, there is potential inconsistency: \{description\}. This warrants caution in the confidence assessment. \\
\bottomrule
\end{tabular}
\end{table*}

\end{document}